\documentclass[11pt,a4paper]{article}
\usepackage[hyperref]{acl2018}
\usepackage{times}
\usepackage{latexsym}
\usepackage{tabularx}
\usepackage{booktabs}
\usepackage{graphicx}
\usepackage{amsmath}

\usepackage{url}

\aclfinalcopy 

\title{Hierarchical Neural Story Generation}

\author{Angela Fan \\\And
  Mike Lewis \\ \\
  Facebook AI Research, Menlo Park\\
  \{angelafan, mikelewis, ynd\}@fb.com \And
  Yann Dauphin \\
  }

\date{}

\begin{document}
\maketitle
\begin{abstract}
We explore \emph{story generation}: creative systems that can build coherent and fluent passages of text about a topic.
We collect a large dataset of 300K human-written stories paired with writing prompts from an online forum.
Our dataset enables hierarchical story generation, where the model first generates a premise, and then transforms it into a passage of text.
We gain further improvements with a novel form of model fusion that improves the relevance of the story to the prompt, and adding a new gated multi-scale self-attention mechanism to model long-range context.
Experiments show large improvements over strong baselines on both automated and human evaluations.
Human judges prefer stories generated by our approach to those from a strong non-hierarchical model by a factor of two to one.
\end{abstract}

\section{Introduction}

Story-telling is on the frontier of current text generation technology: stories must remain thematically consistent across the complete document, requiring modeling very long range dependencies; stories require creativity; and stories need a high level plot, necessitating planning ahead rather than word-by-word generation \cite{wiseman2017challenges}.

We tackle the challenges of story-telling with a hierarchical model, which first generates a sentence called the \emph{prompt} describing the topic for the story, and then conditions on this prompt when generating the story.
Conditioning on the prompt or premise makes it easier to generate consistent stories because they provide grounding for the overall plot. It also reduces the tendency of standard sequence models to drift off topic.

We find that standard sequence-to-sequence (seq2seq) models \cite{sutskever14} applied to hierarchical story generation are prone to degenerating into language models that pay little attention to the writing prompt (a problem that has been noted in other domains, such as dialogue response generation \citep{li:2015}).
This failure is due to the complex and underspecified dependencies between the prompt and the story, which are much harder to model than the closer dependencies required for language modeling (for example, consider the subtle relationship between the first sentence and prompt in Figure \ref{tbl:wp_example}). 

To improve the relevance of the generated story to its prompt, we introduce a fusion mechanism \cite{sriram2017} where our model is trained on top of an pre-trained seq2seq model. To improve over the pre-trained model, the second model must focus on  the link between the prompt and the story. 
For the first time, we show that fusion mechanisms can help seq2seq models build dependencies between their input and output.

\begin{figure}[t]
  \centering \small
  \begin{tabularx}{\linewidth}{X}
    \textbf{Prompt: }The Mage, the Warrior, and the Priest \\
    \toprule
       \textbf{Story:} A light breeze swept the ground, and carried with it still the distant scents of dust and time-worn stone. The Warrior led the way, heaving her mass of armour and muscle over the uneven terrain. She soon crested the last of the low embankments, which still bore the unmistakable fingerprints of haste and fear. She lifted herself up onto the top the rise, and looked out at the scene before her. [...] \\ 
	 \midrule
\end{tabularx}
   \caption{Example prompt and beginning of a story from our dataset. We train a hierarchical model that first generates a prompt, and then conditions on the prompt when generating a story.}
 \label{tbl:wp_example}
\end{figure}

Another major challenge in story generation is the inefficiency of modeling long documents with standard recurrent architectures---stories contain 734 words on average in our dataset.
We improve efficiency using a convolutional architecture, allowing whole stories to be encoded in parallel.
Existing convolutional architectures only encode a bounded amount of context \citep{dauphin17}, so we introduce a novel gated self-attention mechanism that allows the model to condition on its previous outputs at different time-scales.

To train our models, we gathered a large dataset of 303,358 human generated stories paired with writing prompts from an online forum. Evaluating free form text is challenging, so we also introduce new evaluation metrics which isolate different aspects of story generation. 

Experiments show that our fusion and self-attention mechanisms improve over existing techniques on both automated and human evaluation measures. Our new dataset and neural architectures allow for models which can creatively generate longer, more consistent and more fluent passages of text. Human judges prefer our hierarchical model's stories twice as often as those of a non-hierarchical baseline.

\section{Writing Prompts Dataset}

We collect a hierarchical story generation dataset\footnote{ \url{www.github.com/pytorch/fairseq}} from Reddit's \textsc{WritingPrompts} forum.\footnote{\url{www.reddit.com/r/WritingPrompts/}} \textsc{WritingPrompts} is a community where online users inspire each other to write by submitting story premises, or prompts, and other users freely respond. Each prompt can have multiple story responses. The prompts have a large diversity of topic, length, and detail. The stories must be at least 30 words, avoid general profanity and inappropriate content, and should be inspired by the prompt (but do not necessarily have to fulfill every requirement). Figure~\ref{tbl:wp_example} shows an example.

\begin{table}[t]
  \centering 
  \begin{tabular}{ l c }\hline
    \# Train Stories & 272,600 \\
    \# Test Stories & 15,138 \\
    \# Validation Stories & 15,620 \\
    \hline 
    \# Prompt Words & 7.7M \\
    \# Story Words & 200M \\
    \hline 
    Average Length of Prompts & 28.4 \\
    Average Length of Stories & 734.5 \\
 \hline
\end{tabular}
   \caption{Statistics of \textsc{WritingPrompts} dataset}
 \label{tbl:wp_dataset}
\end{table}

We scraped three years of prompts and their associated stories using the official Reddit API. We clean the dataset by removing automated bot posts, deleted posts, special announcements, comments from moderators, and stories shorter than 30 words.
We use \textsc{NLTK} for tokenization. The dataset models full text to generate immediately human-readable stories. 
We reserve 5\% of the prompts for a validation set and 5\% for a test set, and present additional statistics about the dataset in Table~\ref{tbl:wp_dataset}. 

For our experiments, we limit the length of the stories to 1000 words maximum and limit the vocabulary size for the prompts and the stories to words appearing more than 10 times each. We model an unknown word token and an end of document token. This leads to a vocabulary size of 19,025 for the prompts and 104,960 for the stories. As the dataset is scraped from an online forum, the number of rare words and misspellings is quite large, so modeling the full vocabulary is challenging and computationally intensive.

\section{Approach}

The challenges of \textsc{WritingPrompts} are primarily in modeling long-range dependencies and conditioning on an abstract, high-level prompt. Recurrent and convolutional networks have successfully modeled sentences \citep{jozefowicz2016exploring,dauphin17}, but accurately modeling several paragraphs is an open problem. While seq2seq networks have strong performance on a variety of problems, we find that they are unable to build stories that accurately reflect the prompts. We will evaluate strategies to address these challenges in the following sections.

\subsection{Hierarchical Story Generation}

High-level structure is integral to good stories, but language models generate on a strictly-word-by-word basis and so cannot explicitly make high-level plans. We introduce the ability to plan by decomposing the generation process into two levels. First, we generate the premise or prompt of the story using the convolutional language model from \citet{dauphin17}. The prompt gives a sketch of the structure of the story. Second, we use a seq2seq model to generate a story that follows the premise. Conditioning on the prompt makes it easier for the story to remain consistent and also have structure at a level beyond single phrases.

\subsection{Efficient Learning with Convolutional Sequence-to-Sequence Model}
The length of stories in our dataset is a challenge for RNNs, which process tokens sequentially.
To transform prompts into stories, we instead build on the convolutional seq2seq model of \citet{gehring17}, which uses deep convolutional networks as the encoder and decoder. Convolutional models are ideally suited to modeling long sequences, because they allow parallelism of computation within the sequence.  In the Conv seq2seq model, the encoder and decoder are connected with attention modules \cite{bahdanau14} that perform a weighted sum of encoder outputs, using attention at each layer of the decoder. 

\begin{figure}
   \centering
   \includegraphics[width=0.35\textwidth]{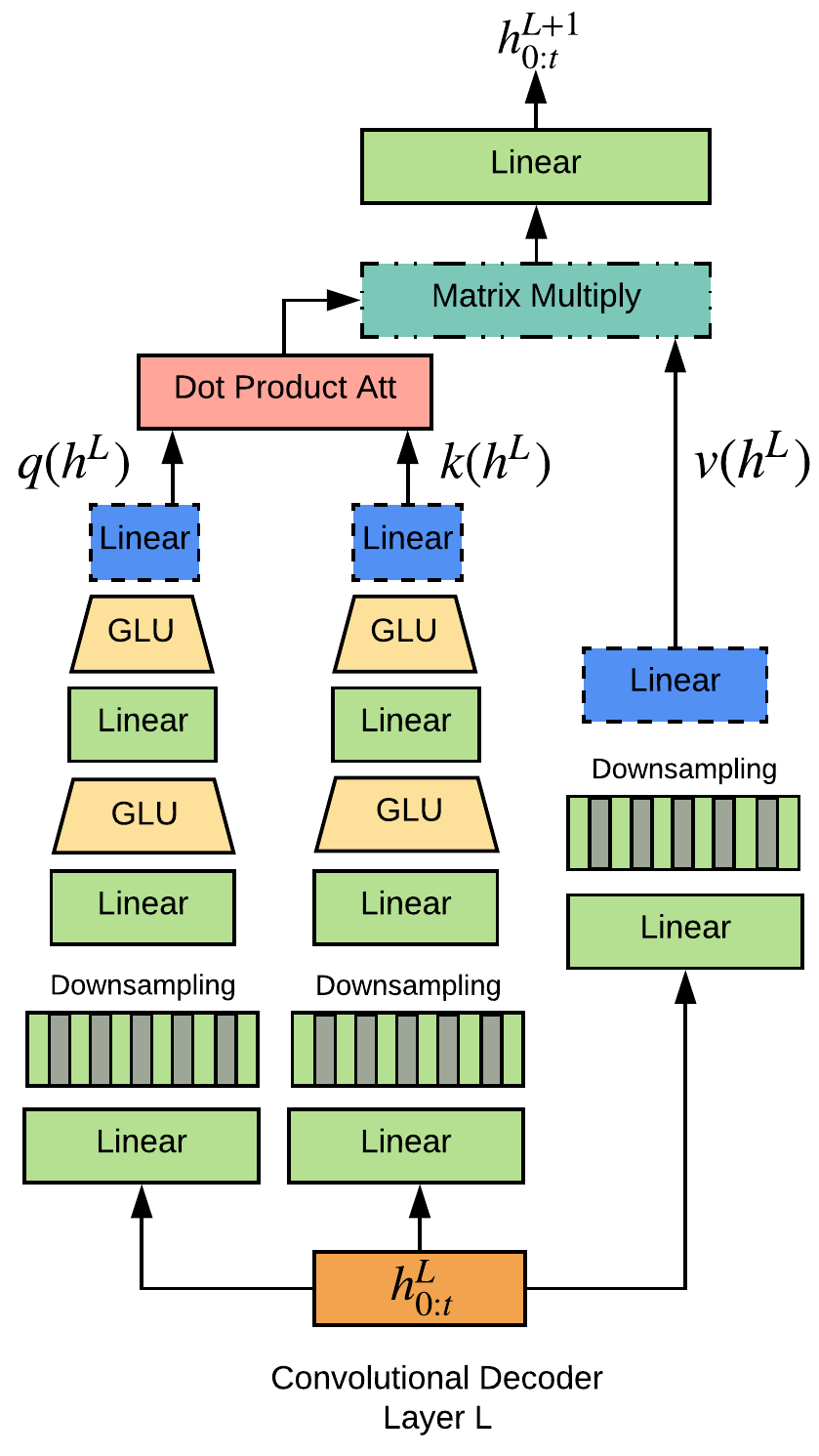}
 \caption{Self-Attention Mechanism of a single head, with GLU gating and downsampling. Multiple heads are concatenated, with each head using a separate downsampling function.}
 \label{fig:multihead_attention}
\end{figure}

\begin{figure}
   \centering
   \includegraphics[width=0.4\textwidth]{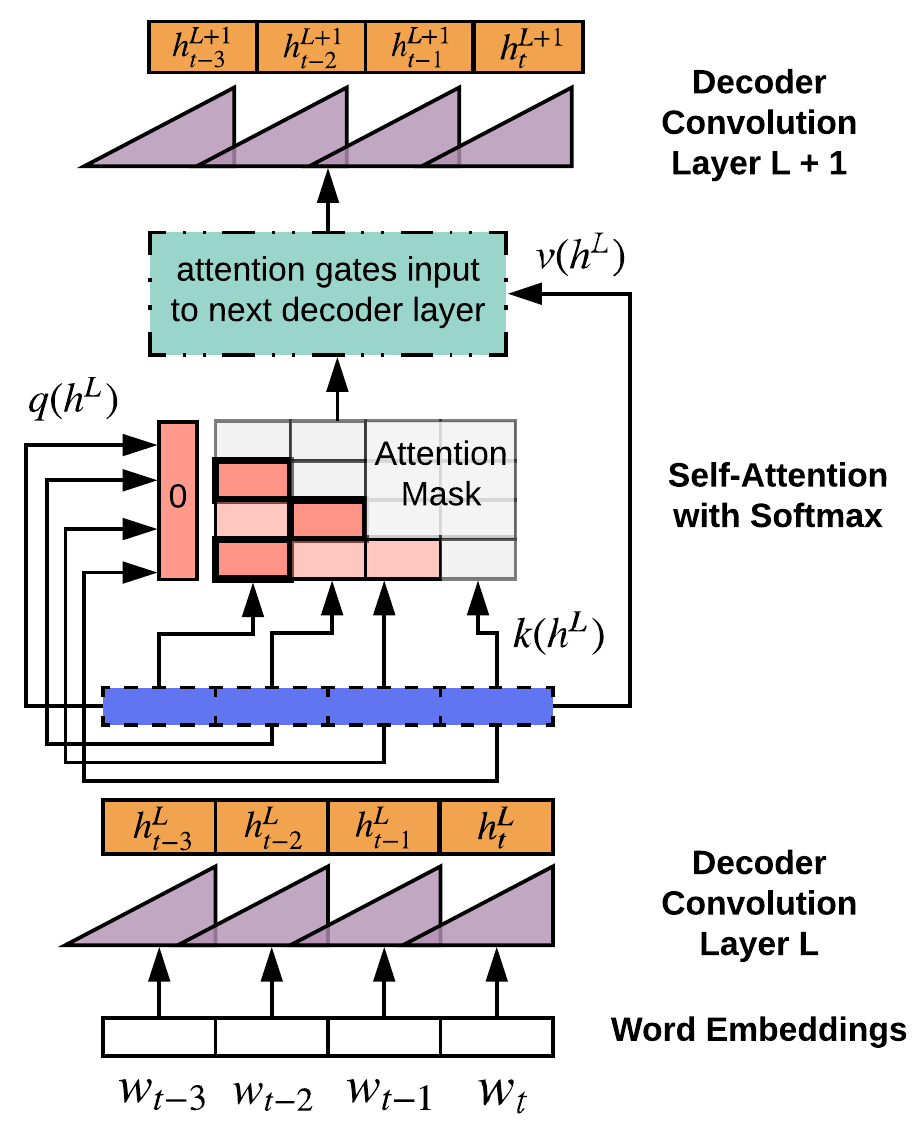}
 \caption{Multihead self-attention mechanism. The decoder layer depicted attends with itself to gate the input of the subsequent decoder layer.}
 \label{fig:attention_figure}
\end{figure}

\subsection{Modeling Unbounded Context with Gated Multi-Scale Self-attention}
CNNs can only model a bounded context window, preventing the modeling of long-range dependencies within the output story.
To enable modeling of unbounded context, we supplement the decoder with a self-attention mechanism \citep{sukhbaatar2015end,vaswani2017}, which allows the model to refer to any previously generated words. The self-attention mechanism improves the model's ability to extract long-range context with limited computational impact due to parallelism. 

\textbf{Gated Attention:} Similar to \citet{vaswani2017}, we use multi-head attention to allow each head to attend to information at different positions. However, the queries, keys and values are not given by linear projections but by more expressive gated deep neural nets with Gated Linear Unit \citep{dauphin17} activations. We show that gating lends the self-attention mechanism crucial capacity to make fine-grained selections. 

\textbf{Multi-Scale Attention:} Further, we propose to have each head operating at a different time scale, depicted in Figure~\ref{fig:multihead_attention}. Thus the input to each head is \emph{downsampled} a different amount---the first head sees the full input, the second every other input timestep, the third every third input timestep,  etc. The different scales encourage the heads to attend to different information. The downsampling operation limits the number of tokens in the attention maps, making them sharper. 

The output of a single attention head is given by
\begin{align}
h^{L+1}_{0:t}=\mathrm{Linear}\Big(& v(h^L_{0:t-1})\\&\odot \mathrm{softmax} (q(h^L_{0:t})k(h^L_{0:t})^\top) \Big) \nonumber
\end{align}
where $h^L_{0:t}$ contains the hidden states up to time $t$ at layer $L$, and  $q,k,v$ are gated downsampling networks as shown in Figure \ref{fig:multihead_attention}. 
Unlike \citet{vaswani2017}, we allow the model to optionally attend to a 0 vector at each timestep, if it chooses to ignore the information of past timesteps (see Figure~\ref{fig:attention_figure}). This mechanism allows the model to recover the non-self-attention architecture and avoid attending to the past if it provides only noise. Additionally, we do not allow the self-attention mechanism to attend to the current timestep, only the past. 

\subsection{Improving Relevance to Input Prompt with Model Fusion}
\label{sect:fusion}

Unlike tasks such as translation, where the semantics of the target are fully specified by the source, the generation of stories from prompts is far more open-ended. We find that seq2seq models ignore the prompt and focus solely on modeling the stories, because the local dependencies required for language modeling are easier to model than the subtle dependencies between prompt and story.

We propose a fusion-based approach to encourage conditioning on the prompt. We train a seq2seq model that has access to the hidden states of a pretrained seq2seq model. Doing so can be seen as a type of boosting or residual learning that allows the second model to focus on what the first model failed to learn---such as conditioning on the prompt. To our knowledge, this paper is the first to show that fusion reduces the problem of seq2seq models degenerating into language models that capture primarily syntactic and grammatical information.

\begin{figure}
   \centering
   \includegraphics[width=0.4\textwidth]{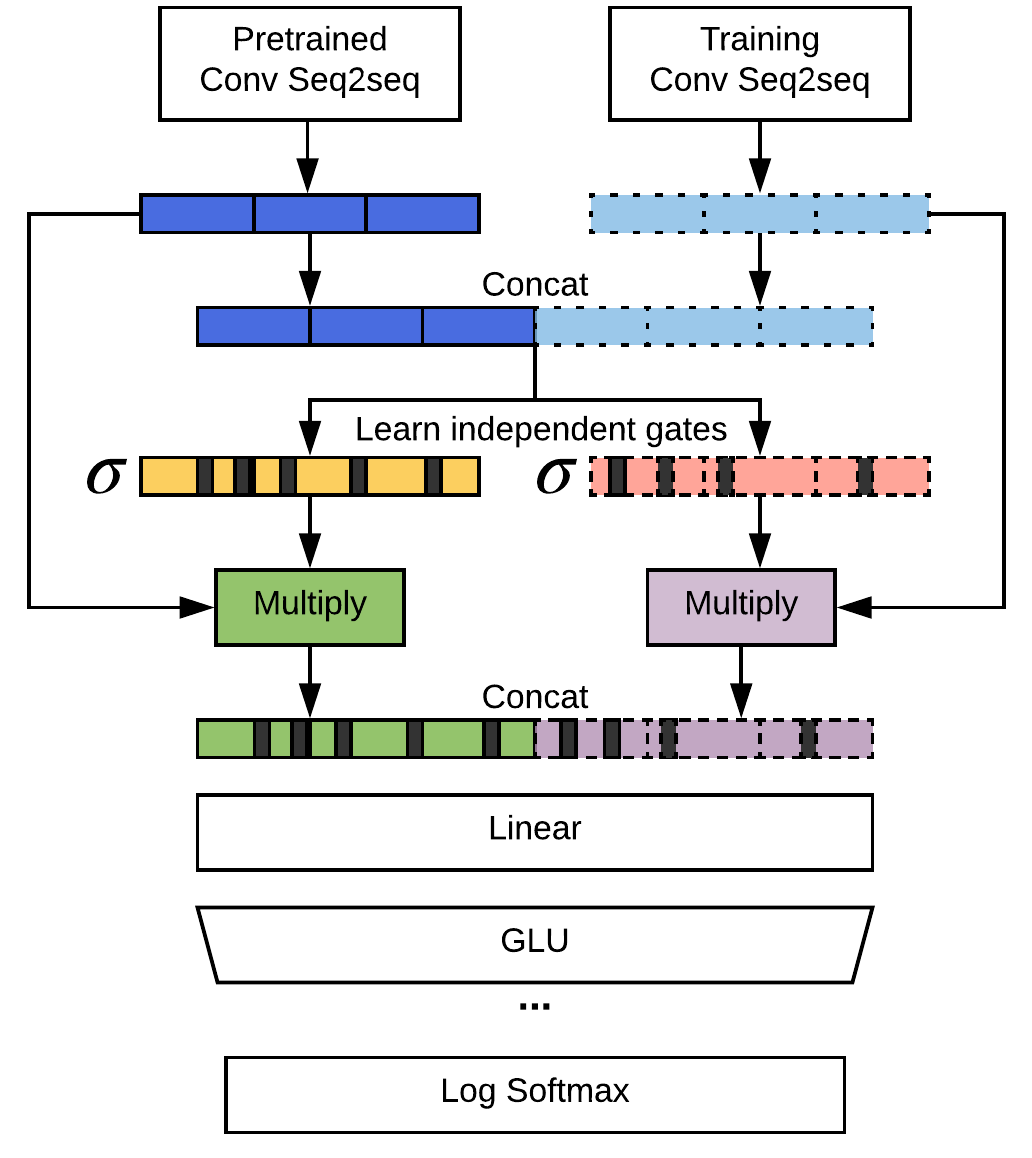}
 \caption{Diagram of our fusion model, which learns a second seq2seq model to improve a pre-trained model. The separate hidden states are combined after gating through concatenation.}
 \label{fig:fusion_model}
\end{figure}

The cold fusion mechanism of \citet{sriram2017} pretrains a language model and subsequently trains a seq2seq model with a gating mechanism that learns to leverage the final hidden layer of the language model during seq2seq training. We modify this approach by combining two seq2seq models as follows (see Figure~\ref{fig:fusion_model}):
\begin{align}
g_t &= \sigma (W [h_t^{\textrm{Training}}; h_t^{\textrm{Pretrained}}] + b )\nonumber\\
h_t &= g_t\circ[h_t^{\textrm{Training}}; h_t^{\textrm{Pretrained}}]\nonumber
\end{align}
where the hidden state of the pretrained seq2seq model and training seq2seq model (represented by $h_t$) are concatenated to learn gates $g_t$. The gates are computed using a linear projection with the weight matrix $W$. The gated hidden layers are combined by concatenation and followed by more fully connected layers with GLU activations (see Appendix). We use layer normalization \citep{ba2016} after each fully connected layer. 

\section{Related Work}

\subsection{Story Generation}

Sequence-to-sequence neural networks \citep{sutskever14} have achieved state of the art performance on a variety of text generation tasks, such as machine translation \citep{sutskever14} and summarization \citep{rush2015neural}. Recent work has applied these models to more open-ended generation tasks, including writing Wikipedia articles \citep{liu2018} and poetry \citep{zhang2014chinese}. 

Previous work on story generation has explored seq2seq RNN architectures \citep{roemmele2016}, but has focused largely on using various content to inspire the stories. For instance, \citet{kiros2015} uses photos to inspire short paragraphs trained on romance novels, and \citet{jain2017} chain a series of independent descriptions together into a short story. \citet{martin2017event} decompose story generation into two steps, first converting text into event representations, then modeling stories as sequences of events before translating back to natural language. Similarly, \citet{harrison2017toward} generate summaries of movies as sequences of events using an RNN, then sample event representations using MCMC. They find this technique can generate text of the desired genre, but the movie plots are not interpretable (as the model outputs events, not raw text). However, we are not aware of previous work that has used hierarchical generation from a textual premise to improve the coherence and structure of stories.

\subsection{Hierarchical Text Generation}

Previous work has proposed decomposing the challenge of generating long sequences of text into a hierarchical generation task. For instance, \citet{li2015hierarchical} use an LSTM to hierarchically learn word, then sentence, then paragraph embeddings, then transform the paragraph embeddings into text. \citet{yarats2017hierarchical} generate a discrete latent variable based on the context, then generates text conditioned upon it. 

\subsection{Fusion Models}

Previous work has investigated the integration of language models with seq2seq models. The two models can be leveraged together without architectural modifications: \citet{ramachandran2016} use language models to initialize the encoder and decoder side of the seq2seq model independently, and \citet{chorowski2016} combine the predictions of the language model and seq2seq model solely at inference time. Recent work has also proposed deeper integration. \citet{gulcehre2015} combined a trained language model with a trained seq2seq model to learn a gating function that joins them. \citet{sriram2017} propose training the seq2seq model given the fixed language model then learning a gate to filter the information from the language model. 

\section{Experimental Setup}

\subsection{Baselines}

We evaluate a number of baselines:

(1) Language Models: Non-hierarchical models for story generation, which do not condition on the prompt. We use both the gated convolutional language (GCNN) model of \citet{dauphin17} and our additional self-attention mechanism. 

(2) seq2seq: using LSTMs and convolutional seq2seq architectures, and Conv seq2seq with decoder self-attention.

(3) Ensemble: an ensemble of two Conv seq2seq with self-attention models.

(4) KNN: we also compare with a KNN model to find the closest prompt in the training set for each prompt in the test set. A TF-IDF vector for each prompt was created using \textsc{fasttext} \citep{bojanowski2016} and \textsc{faiss} \citep{faiss} was used for KNN search. The retrieved story from the training set is limited to 150 words to match the length of generated stories. 

\begin{table}[t]
  \centering 
  \begin{tabular}{ l p{18mm} p{18mm} }\hline
    \bf{Model} & \bf{Valid \newline Perplexity} & \bf{Test \newline Perplexity}\\ \hline\hline
    Conv seq2seq & 45.27 & 45.54 \\
    + self-attention & 42.01 & 42.32 \\
    + multihead & 40.12 & 40.39 \\
    + multiscale & 38.76 & 38.91 \\
    + gating & \textbf{37.37} & \textbf{37.94} \\
 \hline
\end{tabular}
   \caption{Effect of new attention mechanism. Gated multi-scale attention significantly improves the perplexity on the \textsc{WritingPrompts} dataset.}
 \label{tbl:attention_ppl}
\end{table}

\subsection{Fusion Training}

To train the fusion model, we first pretrain a Conv seq2seq with self-attention model on the \textsc{WritingPrompts} dataset. This pretrained model is fixed and provided to the second Conv seq2seq with self-attention model during training time. The two models are integrated with the fusion mechanism described in Section~\ref{sect:fusion}.

\subsection{Training}

We implement models with the fairseq-py library in PyTorch. Similar to \citet{gehring17}, we train using the Nesterov accelerated gradient method \citep{sutskever13} using gradient clipping \citep{pascanu13}. We perform hyperparameter optimization on each of our models by cross-validating with random search on a validation set. We provide model architectures in the appendix. 

\begin{table*}[!t]
  \centering 
  \begin{tabular}{ l c c c }\hline
    \bf{Model} & \bf{\# Parameters (mil)} & \bf{Valid Perplexity} & \bf{Test Perplexity}\\ \hline\hline
    GCNN LM & 123.4 & 54.50 & 54.79 \\
    GCNN + self-attention LM & 126.4 & 51.84 & 51.18 \\
    \hline
    LSTM seq2seq & 110.3 & 46.83 & 46.79 \\
    Conv seq2seq & 113.0 & 45.27 & 45.54 \\
    Conv seq2seq + self-attention & 134.7 & 37.37 & 37.94 \\
    Ensemble: Conv seq2seq + self-attention & 270.3 & 36.63 & 36.93 \\
    Fusion: Conv seq2seq + self-attention & 255.4 & \textbf{36.08} & \textbf{36.56} \\
 \hline
\end{tabular}
   \caption{Perplexity on \textsc{WritingPrompts}. We  dramatically improve over standard seq2seq models.}
 \label{tbl:ppl}
\end{table*}

\subsection{Generation}

We generate stories from our models using a \emph{top-k random sampling} scheme. At each timestep, the model generates the probability of each word in the vocabulary being the likely next word. We randomly sample from the $k = 10$ most likely candidates from this distribution. Then, subsequent timesteps generate words based on the previously selected words. We find this sampling strategy substantially more effective than beam search, which tends to produce common phrases and repetitive text from the training set \cite{vijayakumar2016diverse, shao2017generating}. Sentences produced by beam search tend to be short and generic. Completely random sampling can introduce very unlikely words, which can damage generation as the model has not seen such mistakes at training time. The restriction of sampling from the 10 most likely candidates reduces the risk of these low-probability samples. 

For each model, we tune a temperature parameter for the softmax at generation time. To ease human evaluation, we generate stories of 150 words and do not generate unknown word tokens.

For prompt generation, we use a self-attentive GCNN language model trained with the same prompt-side vocabulary as the sequence-to-sequence story generation models. The language model to generate prompts has a validation perplexity of  63.06. Prompt generation is conducted using the top-k random sampling from the 10 most likely candidates, and the prompt is completed when the language model generates the end of prompt token.

\begin{figure}[!t]
   \centering
   \includegraphics[width=0.5\textwidth]{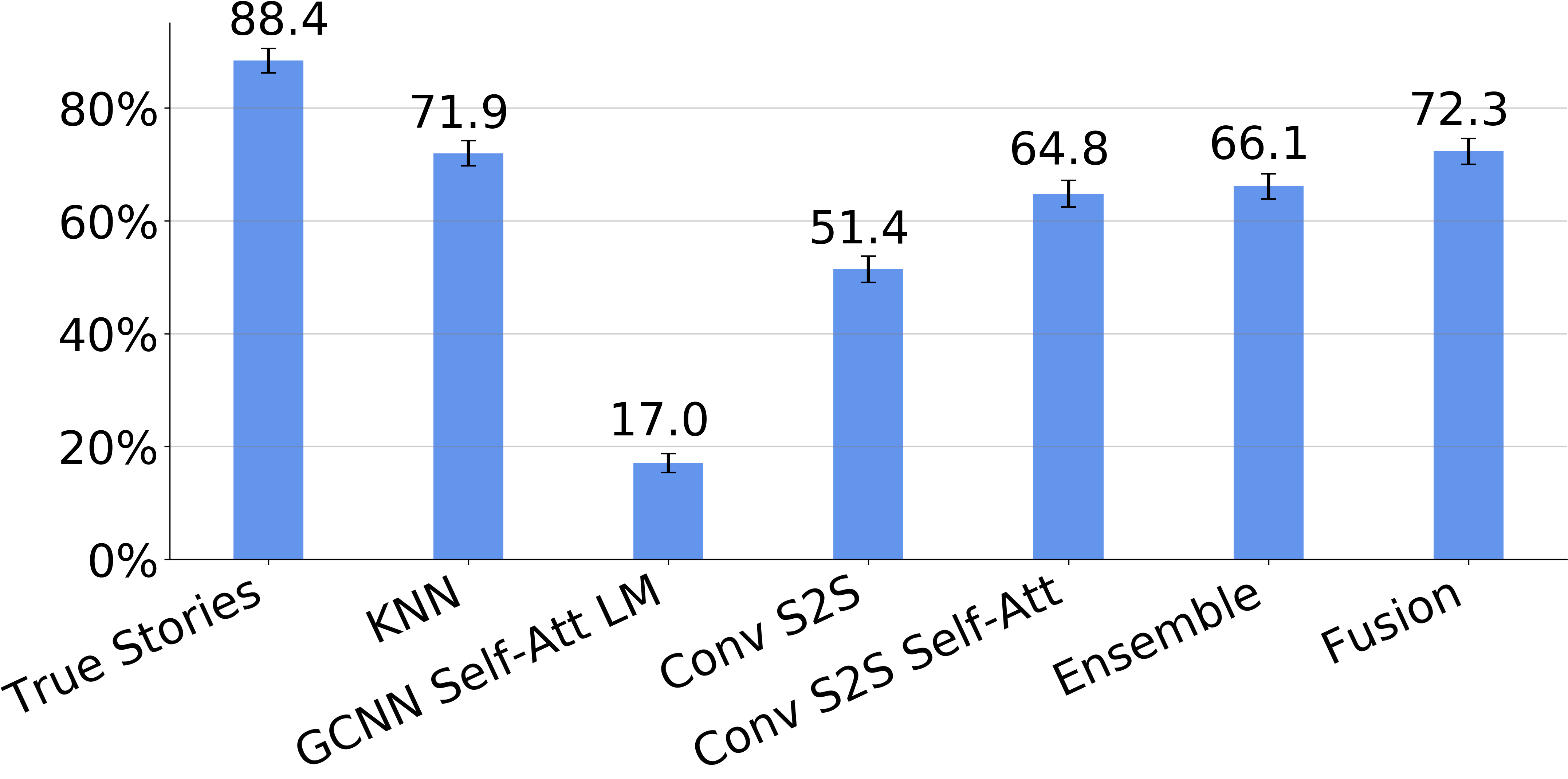}
 \caption{Human accuracy at pairing stories with the prompts used to generate them. People find that our fusion model significantly improves the link between the prompt and generated stories. 
 }
 \label{fig:pairing_task}
\end{figure}

\begin{figure}[!t]
   \centering
   \includegraphics[width=.5\textwidth]{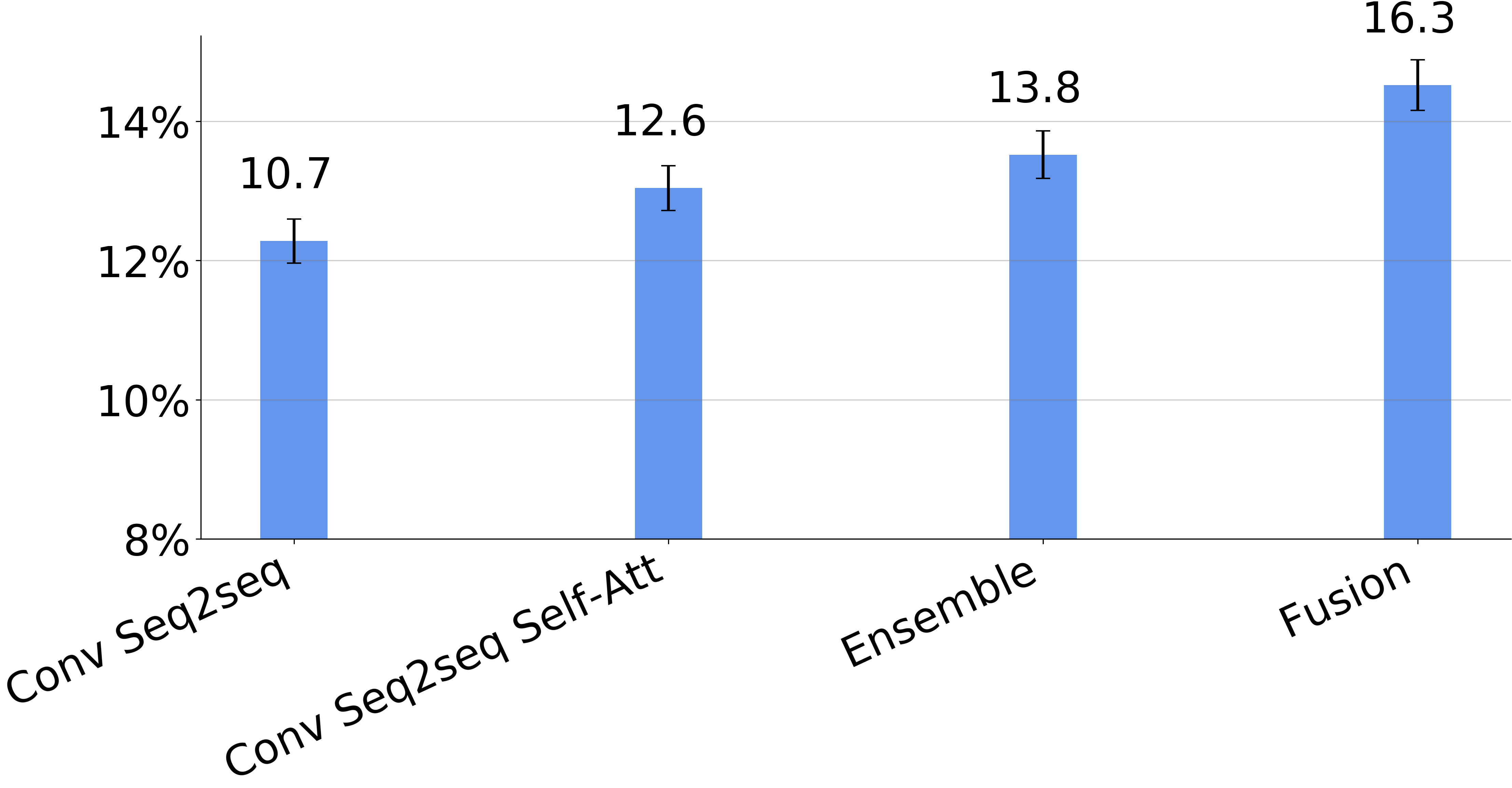}
 \caption{Accuracy of prompt ranking. The fusion model most accurately pairs prompt and stories.
 }
 \label{fig:likelihood_ranking}
\end{figure}

\begin{figure}[!t]
   \centering
   \includegraphics[width=.5\textwidth]{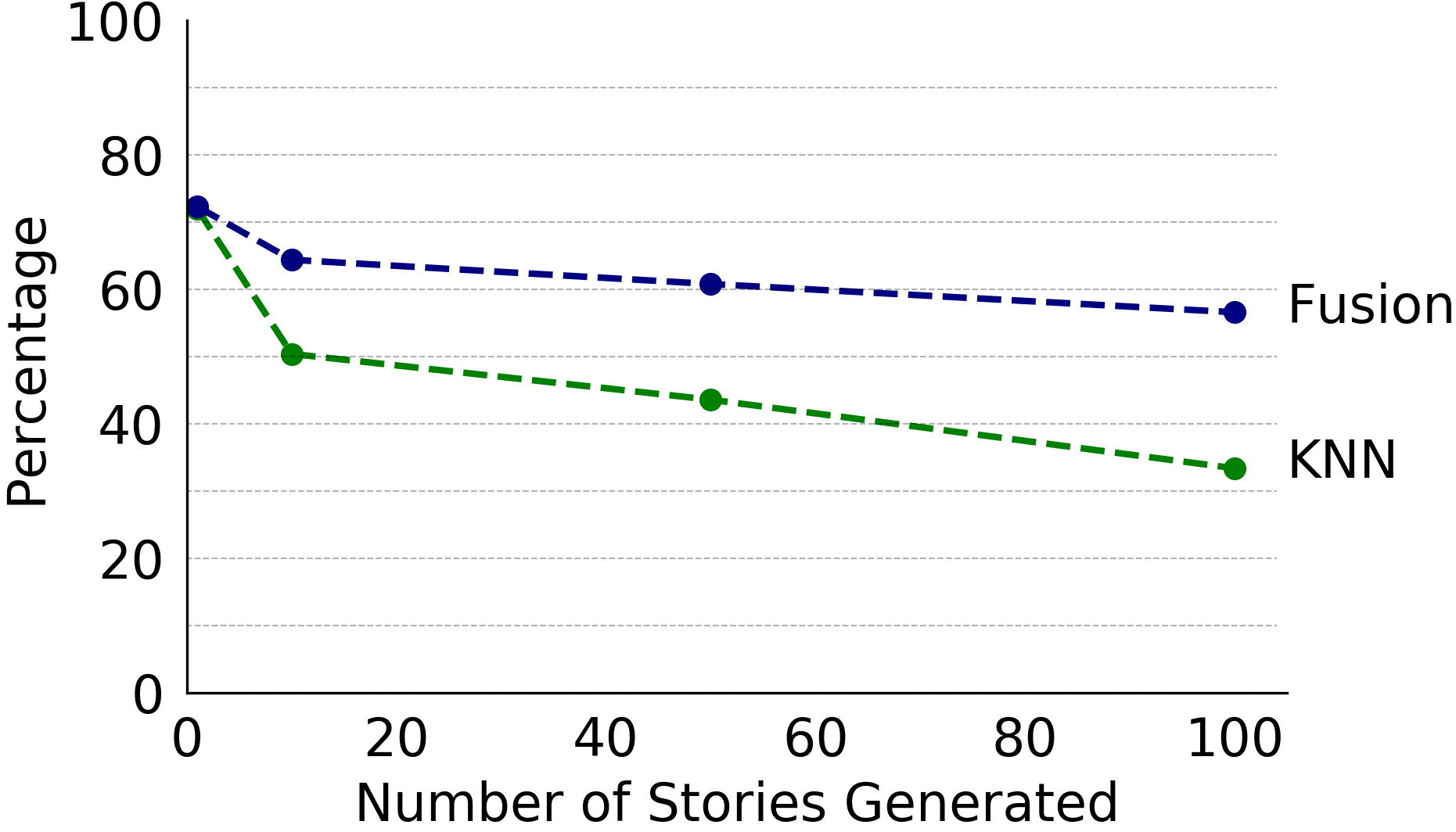}
 \caption{Accuracy on the prompt/story pairing task vs. number of generated stories. Our generative fusion model can produce many stories without degraded performance, while the KNN can only produce a limited number relevant stories.}
 \label{fig:knn_v_fusion}
\end{figure}

\begin{table}[!t]
  \centering 
  \begin{tabular}{ l c }\hline
    \bf{Model} & \bf{Human}\\ 
    & {\bf Preference}\\
    \hline\hline
    Language model &  32.68\% \\
    Hierarchical Model & \textbf{67.32\%} \\
 \hline
\end{tabular}
   \caption{Effect of Hierarchical Generation. Human judges prefer stories that were generated hierarchically by first creating a premise and creating a full story based on it with a seq2seq model.}
 \label{tbl:hierarchical_generation}
\end{table}

\begin{figure*}
   \centering
   \includegraphics[width=.95\textwidth]{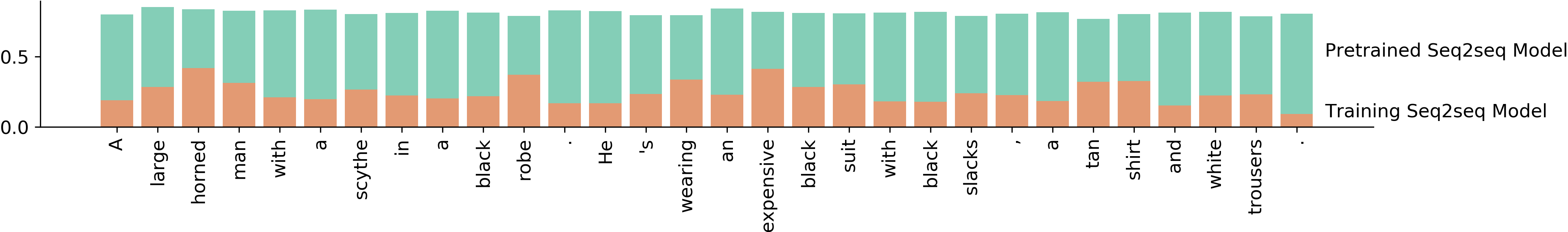}
 \caption{Average weighting of each model in our Fusion model for the beginning of the generated story for the prompt \emph{Gates of Hell}. The fused model (orange) is primarily used for words which are closely related to the prompt, whereas generic words are generated by the pre-trained model (green).
 }
 \label{fig:gate_comparison}
\end{figure*}

\subsection{Evaluation}

We propose a number of evaluation metrics to quantify the performance of our models. Many commonly used metrics, such as \textsc{BLEU} for machine translation or \textsc{ROUGE} for summarization, compute an n-gram overlap between the generated text and the human text---however, in our open-ended generation setting, these are not useful. We do not aim to generate a specific story; we want to generate viable and novel stories. We focus on measuring both the fluency of our models and their ability to adhere to the prompt.  

For automatic evaluation, we measure \emph{model perplexity} on the test set and \emph{prompt ranking accuracy}. Perplexity is commonly used to evaluate the quality of language models, and it reflects how fluently the model can produce the correct next word given the preceding words. We use prompt ranking to assess how strongly a model's output depends on its input. Stories are decoded under 10 different prompts---9 randomly sampled prompts and 1 true corresponding prompt---and the likelihood of the story given the various prompts is recorded. We measure the percentage of cases where the true prompt is the most likely to generate the story. In our evaluation, we examined 1000 stories from the test set for each model.

For human evaluation, we use Amazon Mechanical Turk to conduct a \emph{triple pairing task}. We use each model to generate stories based on held-out prompts from the test set. Then, groups of three stories are presented to the human judges. The stories and their corresponding prompts are shuffled, and human evaluators are asked to select the correct pairing for all three prompts. 105 stories per model are grouped into questions, and each question is evaluated by 15 judges.

Lastly, we conduct human evaluation to evaluate the importance of \emph{hierarchical generation} for story writing. We use Amazon Mechanical Turk to compare the stories from hierarchical generation from a prompt with generation without a prompt. 400 pairs of stories were evaluated by 5 judges each in a blind test. 

\section{Results}
We analyze the effect of our modeling improvements on the \textsc{WritingPrompts} dataset.

\paragraph{Effect of Hierarchical Generation:}  We explore leveraging our dataset to perform hierarchical story generation by first using a self-attentive GCNN language model to generate a prompt, and then using a fusion model to write a story given the generated prompt. We evaluate the effect of hierarchical generation using a human study in Table~\ref{tbl:hierarchical_generation}. 400 stories were generated from a self-attentive GCNN language model, and another 400 were generated from our hierarchical fusion model given generated prompts from a language model. In a blind comparison where raters were asked to choose the story they preferred reading, human raters preferred the hierarchical model 67\% of the time. 

\paragraph{Effect of new attention mechanism:} Table~\ref{tbl:attention_ppl} shows the effect of the proposed additions to the self-attention mechanism proposed by \citet{vaswani2017}. Table \ref{tbl:ppl} shows that deep multi-scale self-attention and fusion each significantly improve the perplexity compared to the baselines. In combination these additions to the Conv seq2seq baseline reduce the perplexity by 9 points.

\paragraph{Effect of model fusion:} Results in Table~\ref{tbl:ppl} show that adding our fusion mechanism substantially improves the likelihood of human-generated stories, and even outperforms an ensemble despite having fewer parameters.
We observe in Figure \ref{fig:pairing_task} that fusion has a much more significant impact on the topicality of the stories.
In comparison, ensembling has no effect on people's ability to associate stories with a prompt, but adding model fusion leads improves the pairing accuracy of the human judges by 7\%.
These results suggest that by training a second model on top of the first, we have encouraged that model to learn the challenging additional dependencies to relate to the source sequence. To our knowledge, these are the first results to show that fusion has such capabilities.

\paragraph{Comparison with Nearest Neighbours:} 
Nearest Neighbour Search (KNN) provides a strong baseline for text generation. 
Figure \ref{fig:pairing_task} shows that the fusion model can match the performance of nearest neighbour search in terms of the connection between the story and prompt.
The real value in our generative approach is that it can produce an unlimited number of stories, whereas KNN can never generalize from its training data.
To quantify this improvement, Figure \ref{fig:knn_v_fusion} plots the relevance of the \emph{k}th best story to a given prompt; the performance of KNN degrades much more rapidly.


\begin{table*}[!t]
  \centering \small
  \begin{tabularx}{\linewidth}{X}
    \textbf{\emph{Generation with Fusion of Conv seq2seq + Self-Attention}} \\
    \toprule
    \textbf{Example Prompt 1: } Aliens start abducting humans \\
    \toprule
       It has been two weeks, and the last of my kind has gone. It is only a matter of time until there will be nothing left. I'm not sure what the hell is going on... I can't think. I can hear a distant scream. I think of a strange, alien sound. I try to speak, but am interrupted by something, something that feels like a drum, I ca not tell. I mean I'm just a little bit older than an average human. But I can, and I can feel the vibrations . I hear the sound of a ship approaching. The ground quakes at the force of the impact, and a loud explosion shatters the silence. \\ 
	 \midrule
\end{tabularx}
~\\
  \begin{tabularx}{\linewidth}{X}
    \textbf{Example Prompt 2: } The scientists have discovered something terrible .  \\
    \toprule
       The scientist stood there, a little dazed as he stared.  
       
       ``What is it?'' He asked. 
       
       ``This...this...Thing...This is a virus. A chemical that can destroy entire planet and it is a very small, complex, chemical that could destroy any planet.'' The scientist replied. His lab assistant looked down at the tablet. 
       
       ``I've just discovered it. I can't believe it. It looks like it's made of some sort of chemical that's very dangerous.'' 
       
       `` Well, there's a virus on the ground. It's very effective...'' 
       `` I can't believe what it is.'' he said, pointing to the scientist . 
       
       `` We don't know what this thing is. We haven't seen anything like it . We can't even see anything like this. '' Dr. Jones stared at the scientist for a moment. 
       
       ``What do you mean what does it do ?'' 
       
       ``It...It 's a monster.'' \\
	 \midrule
\end{tabularx}
~\\
  \begin{tabularx}{\linewidth}{X}
    \textbf{Example Prompt 3: } The silence before the storm comes  \\
    \toprule
        ``I'm sorry,'' she said ``I can't do this anymore.'' 
        
        She sat in the grass, the grass blowing behind her as she sat staring at the sky. Her eyes were still closed but she was still staring at it. Her hand was still shaking, her eyes closed as she watched the rain roll in and out of view. She could see the world through her eyelids now. Her hands were still shaking like she hadn't eaten in days. Her fingers were still clenched tight and her fingers were now shaking. She wasn't even sure if she was going to cry or if she just wanted to cry. It wasn't until she was close enough to hear the words. 
        
        ``I ...I just...I need to...I ...'' She was crying now and then, the rain was picking up. Her lips were shaking now and she could see that the storm was getting closer and closer. Her heart stopped when she heard the thunder. She knew it was coming. The storm was \\
	 \midrule
\end{tabularx}
~\\~\\
  \begin{tabularx}{\linewidth}{X}
  \textbf{\emph{Generation with Language Model + Self-Attention Baseline}} \\
    \toprule
    \textbf{Example 1: }\\
    \toprule
       I've always been the curious type. Growing up in the ghetto, I've been around for more than a year now. I still haven't heard it on my own. I'm not sure if I'm a good person. But I'm not the only one. I've been to the hospital, but it's not my first time. I've been to my psychiatrist's journals before. ``Hey, are you ... are you okay?'' I asked. ``Yes'' I said. ``What's wrong?'' she replied, ``I was just ... Doing something.'' She said, ``I've never been a bad person.'' \\ 
	 \midrule
\end{tabularx}
~\\
  \begin{tabularx}{\linewidth}{X}
    \textbf{Example 2: }\\
    \toprule
    The man was an accountant. He had to be. He had to be the next president. I looked back over the top and saw that his wife was crying in the kitchen. I looked at the clock. It seemed to be coming slower, but I knew if I did it would not be long before I was in my own home. I wasn't sure. I had a hard time finding the right words to say. I was about to leave when he suddenly became angry and began talking to me. ``Hello, sir, I'm John. What is your name?'' ``My name is Manuel and I'm a journalist.'' I said \\
	 \midrule
\end{tabularx}
   \caption{Example stories generated by the proposed hierarchical fusion approach compared to stories generated by a language model. Stories generated by the fusion model relate to the desired prompt and show increased coherence between sentences and ability to stay on one topic compared to the language modeling baseline.}
 \label{tbl:example_stories}
\end{table*}

\section{Discussion}

\subsection{Generation Quality}

Our proposed fusion model is capable of generating unique text without copying directly from the training set. When analyzing 500 150-word generated stories from test-set prompts, the average longest common subsequence is 8.9. In contrast, the baseline Conv seq2seq model copies 10.2 words on average and the KNN baseline copies all 150 words from a story in the training set. 

Figure~\ref{fig:gate_comparison} shows the values of the fusion gates for an example story, averaged at each timestep. The pretrained seq2seq model acts similarly to a language model  producing common words and punctuation. The second seq2seq model learns to focus on rare words, such as \emph{horned} and \emph{robe}.

However, the fusion model has limitations. Using random sampling to generate can produce errors. For example, \emph{can't} is tokenized to \emph{ca} \emph{n't}, and the model occasionally produces the first token but misses the second. A similar error is after one line of dialogue, the model may move to another line of dialogue without generating a newline token. A further obstacle is repetition. The model focuses frequently on what it has recently produced, which leads to the generation of similar text multiple times. 

In the generation of prompts using the GCNN language model, we find that prompts are fairly generic compared to human prompts. Language models often struggle to model rare words accurately, as the probability distribution over the next word is dominated by more common words. This tends to produce similar prompts, particularly at the start --- we see many prompts that start with \emph{the man}. In contrast, many of the human prompts are very unique (e.g. prompting stories in fantasy worlds such as Harry Potter and Game of Thrones) and the language model rarely produces the specific vocabulary required by these settings.   

\subsection{Use of Attention}

We analyze the encoder-decoder attention in the fusion model and find that unlike attention maps in machine translation, where each decoder timestep tends to attend to a different word on the encoder-side, the attention map for each decoder timestep looks similar and focuses mainly on salient words in the prompt. We further look at the usage of the self-attention layers within the decoder. While they could be leveraged to look at words generated very far in the past, at many timesteps the self-attention focuses on the recent past. 

\section{Conclusion}

We have collected the first dataset for creative text generation based on short writing prompts. This new dataset pushes the boundaries of text generation by requiring longer range dependencies and conditioning on an abstract premise. Building on this dataset, we show through automatic and human evaluation that novel hierarchical models, self-attention mechanisms and model fusion significantly improves the fluency, topicality, and overall quality of the generated stories.

\bibliographystyle{acl_natbib}
\bibliography{acl2018}

\section{Appendix of Model Architectures}

\subsection{GCNN Language Model + Self-Attention}

9 layers with hidden unit sizes $512 \times 4, 768 \times 2, 1024 \times 3$ and convolutional kernel widths $4 \times 2, 1, 4 \times 3, 1, 3 \times 2$. Learning rate 1, momentum 0.99, dropout 0.1, embedding size 300, l2 normalization $1e^{-7}$, 4 decoder self-attention heads.

\subsection{Conv seq2seq + self-attention}
\label{sect:seq2seq_architecture}

3 layers in encoder with hidden unit sizes $128 \times 2, 512$ and convolutional kernel widths $3 \times 3$. 8 layers in the decoder with hidden unit sizes $512 \times 4, 768 \times 2, 1024$ with convolutional kernel widths $4 \times 8$. Learning rate 0.25, momentum 0.99, dropout 0.3, embedding size 256, output embedding size 256, l2 nomalization $1e^{-7}$, 4 decoder self-attention heads.

\subsection{Ensemble: Conv seq2seq + self-attention}

Two different Conv seq2seq models were trained and ensembled together by averaging with equal weights.

\subsection{Fusion: Conv seq2seq + self-attention}

The pretrained seq2seq model is the model in Section~\ref{sect:seq2seq_architecture}. The additional fused model has the following architecture:

5 layers in the encoder with hidden unit sizes $128 \times 2, 512 \times 3$ and convolutional kernel widths $3 \times 5$. 5 layers in the decoder with hidden unit sizes $512 \times 3, 768 \times 2$ and convolutional kernel widths $4 \times 5$. Learning rate 0.25, momentum 0.99, dropout 0.3, embedding size 256, output embedding size 256, l2 normalization $1e^{-7}$, 4 decoder self-attention heads.

\end{document}